\documentclass[11pt,a4paper]{article}
\usepackage[hyperref]{acl2018}
\usepackage{times}
\usepackage{latexsym}
\usepackage{subfig}

\usepackage{array}
\usepackage{colortbl}

\usepackage{url}

\usepackage{pgfplots}
\pgfplotsset{width=10cm,compat=1.9} 
\definecolor{darkblue}{rgb}{0,0,.5}
\definecolor{darkgreen}{rgb}{0,.5,0}
\definecolor{lightgray}{rgb}{.8,.8,.8}

\aclfinalcopy 
\def\aclpaperid{***} 

\title{On the Impact of Various Types of Noise on Neural Machine Translation}

\author{Huda Khayrallah \qquad \qquad\\
  Center for Language \& Speech Processing \qquad\qquad\\
  Computer Science Department \qquad\qquad \\
  Johns Hopkins University \qquad\qquad \\
  {\tt huda@jhu.edu}\qquad\qquad \\
  \And
 \qquad\qquad Philipp Koehn\\
\qquad\qquad Center for Language \& Speech Processing \\
 \qquad\qquad Computer Science Department\\
 \qquad\qquad Johns Hopkins University\\
 \qquad\qquad {\tt phi@jhu.edu}}

\begin{document}
\maketitle
\begin{abstract}
We examine how various types of noise in the parallel training data impact the quality of neural machine translation systems. We create five types of artificial noise and analyze how they degrade performance in neural and statistical machine translation. We find that neural models are generally more harmed by noise than statistical models. For one especially egregious type of noise they learn to just copy the input sentence.
\end{abstract}

\section{Introduction}
While neural machine translation (NMT) has shown large gains in quality over statistical machine translation (SMT) \cite{wmtresults}, there are significant exceptions to this, such as low resource and domain mismatch data conditions \cite{koehn-knowles:2017:NMT}.

In this work, we consider another challenge to neural machine translation: noisy parallel data. 
As a motivating example, consider the numbers in Table~\ref{tab:paracrawl}. Here, we add an equally sized noisy web crawled corpus to high quality training data provided by the shared  task of the Conference on Machine Translation (WMT). This addition leads to a 1.2 BLEU point increase for the statistical machine translation system, but degrades the neural machine translation system by 9.9 BLEU. 

The maxim {\em more data is better} that holds true for statistical machine translation does seem to come with some caveats for neural machine translation. The added data cannot be too noisy.
But what kind of noise harms neural machine translation models? 

In this paper, we explore several types of noise and assess their impact by adding synthetic noise to an existing parallel corpus. We find that for almost all types of noise, neural machine translation systems are harmed more than statistical machine translation systems. We discovered that one type of noise, copied source language segments, has a catastrophic impact on neural machine translation quality, leading it to learn a copying behavior that it then exceedingly applies. 

\section{Related Work} 

\begin{table}
\begin{center}
\begin{tabular}{l|l|l}
& \bf NMT & \bf SMT \\ \hline
WMT17 & 27.2 & 24.0 \\
+ noisy corpus & 17.3 (--9.9) & 25.2 (+1.2) \\ \hline
\end{tabular}
\end{center}
\caption{
Adding noisy web crawled data (raw data from {\tt \small paracrawl.eu}) to a WMT 2017 German--English statistical system obtains small gains (+1.2 BLEU), a neural system falls apart (--9.9 BLEU).
}
\label{tab:paracrawl}
\end{table}

There is a robust body of work on filtering out noise in parallel data. For example: \citet{MTS-2011-Taghipour} use an outlier detection algorithm to filter a parallel corpus; \citet{D17-1318} generate synthetic noisy data (inadequate and non-fluent translations) and use this data to train a classifier to identify  good sentence pairs from a noisy corpus; and \citet{cui-EtAl:2013:Short} use a graph-based random walk algorithm and extract phrase pair scores to weight the phrase translation probabilities to bias towards more trustworthy ones. 

Most of this work was done in the context of statistical machine translation, but more recent work \citep{carpuat-vyas-niu:2017:NMT} targets neural models. That work focuses on identifying semantic differences in translation pairs using cross-lingual textual entailment and additional length-based features, and demonstrates that removing such sentences improves neural machine translation performance. 

As \citet{MTS-2011-Rarrick} point out, one problem of parallel corpora extracted from the web is translations that have been created by machine translation. \citet{venugopal-EtAl:2011:EMNLP} propose a method to watermark the output of machine translation systems to aid this distinction.  \citet{antonova-misyurev:2011:BUCC} report that rule-based machine translation output can be detected due to certain word choices, and statistical machine translation output due to lack of reordering. 

In 2016, a shared task on sentence pair filtering was organized\footnote{NLP4TM 2016:  \href{ http://rgcl.wlv.ac.uk/nlp4tm2016/shared-task}{ rgcl.wlv.ac.uk/nlp4tm2016/shared-task}} \cite{Barbu2016}, albeit in the context of cleaning translation memories which tend to be cleaner than web crawled data. This year, a shared task is planned for the type of noise that we examine in this paper.\footnote{\href{http://statmt.org/wmt18/parallel-corpus-filtering.html}{statmt.org/wmt18/parallel-corpus-filtering.html}}

\citet{Belinkov:Bisk:noise} investigate noise in neural machine translation, but they focus on creating systems that can {\em translate}  the kinds of orthographic errors (typos, misspellings, etc.) that humans can comprehend. In contrast, we address noisy {\em training} data and focus on types of noise occurring in web-crawled corpora.

There is a rich literature on data selection which aims at sub-sampling parallel data relevant for a task-specific machine translation system \citep{axelrod-he-gao:2011:EMNLP}. \citet{D17-1148} find that the existing data selection methods developed for statistical machine translation  are less effective for neural machine translation. This is different from our goals of handling noise since those methods tend to discard perfectly fine sentence pairs (say, about cooking recipes) that are just not relevant for the targeted domain (say, software manuals). Our work is focused on noise that is harmful for all domains.  

Since we begin with a clean parallel corpus and potentially noisy data to it, this work can be seen as a type of data augmentation. \citet{sennrich-haddow-birch:2016:P16-11} incorporate monolingual corpora into NMT by first translating it using an NMT system trained in the opposite direction. While such a corpus has the potential to be noisy, the method is very effective.  \citet{currey-micelibarone-heafield:2017:WMT} create additional parallel corpora by copying monolingual corpora in the target language into the source, and find it improves over back-translation for some language pairs. \citet{fadaee-bisazza-monz:2017:Short2} improve NMT performance in low-resource settings by altering existing sentences to create training data that includes rare words in different contexts.

\section{Real-World Noise}
What types of noise are prevalent in crawled web data? We manually examined 200 sentence pairs of the above-mentioned Paracrawl corpus and classified them into several error categories. Obviously, the results of such a study depend very much on how crawling and extraction is executed, but the results (see Table~\ref{tab:paracrawl-noise}) give some indication of what noise to expect.

We classified any pairs of German and English sentences that are not translations of each other as misaligned sentences. These may be caused by any problem in alignment processes (at the document level or the sentence level), or by forcing the alignment of content that is not indeed parallel. Such misaligned sentences are the biggest source of error (41\%).

There are three types of wrong language content (totaling 23\%): one or both sentences may be in a language different from German and English (3\%), both sentences may be German (10\%), or both languages may be English (10\%).

4\% of sentence pairs are untranslated, i.e., source and target are identical. 2\% sentence pairs consist of random byte sequences, only HTML markup, or Javascript. A number of sentence pairs have very short German or English sentences, containing at most 2 tokens (1\%) or 5 tokens (5\%).

\begin{table}
\begin{center}
\begin{tabular}{l|r}
\bf Type of Noise & \bf Count\\ \hline
Okay & 23\%\\ \hline
Misaligned sentences & 41\%\\\hline
Third language & 3\% \\
Both English & 10\% \\
Both German & 10\% \\\hline
Untranslated sentences & 4\%\\ \hline
Short segments ($\leq$2 tokens) & 1\% \\
Short segments (3--5 tokens) & 5\% \\\hline
Non-linguistic characters & 2\% \\\hline
\end{tabular}
\caption{Noise in the raw Paracrawl corpus.}
\label{tab:paracrawl-noise}
\end{center}
\end{table}

Since it is a very subjective value judgment what constitutes disfluent language, we do not classify these as errors.  However, consider the following sentence pairs that 
we did count as okay, although they contain mostly untranslated names and numbers.
\vspace{2mm}

\noindent {\em 
\begin{tabular}{p{7.2cm}} \hline
DE: Anonym 2 24.03.2010 um 20:55 314 Kommentare\\
EN: Anonymous 2 2010-03-24 at 20:55 314 Comments\\ \hline
DE: \&lt; \&lt; erste \&lt; zur\"uck Seite 3 mehr letzte \&gt; \&gt;\\
EN: \&lt; \&lt; first \&lt; prev. page 3 next last \&gt; \&gt;\\\hline
\end{tabular}}
\vspace{2mm}

At first sight, some types of noise seem to be easier to automatically identify than others. However, consider, for instance, content in a wrong language. While there are established methods for language identification (typically based on character n-grams), these do not work well on a sentence-level basis, especially for short sentences. Or, take the apparently obvious problem of untranslated sentences. If they are completely identical, that is easy to spot --- although even those may have value, such as the list of country names which are often spelled identical in different languages. However, there are many degrees of near-identical content of unclear utility.

\section{Types of Noise} 
The goal of this paper is not to develop methods to detect noise but to ascertain the impact of different types of noise on translation quality when present in parallel data. We hope that our findings inform future work on parallel corpus cleaning.

We now formally define five types of naturally occurring noise and describe how we simulate them. By creating artificial noisy data, we avoid the hard problem of detecting specific types of noise but are still able to study their impact.

\paragraph{\textsc{Misaligned Sentences}}
As shown above, a common source of noise in parallel corpora is faulty document or sentence alignment. This results in sentences that are not matched to their translation. Such noise is rare in corpora such as Europarl where strong clues about debate topics and speaker turns reduce the scale of the task of alignment to paragraphs, but more common in the alignment of less structured web sites.
We artificially create misaligned sentence data by randomly shuffling the order of sentences on one side of the original clean parallel training corpus.

\paragraph{\textsc{Misordered Words}}
Language may be disfluent in many ways. This may be the product of machine translation, poor human translation, or heavily specialized language use, such as bullet points in product descriptions (recall also the examples above). We consider one extreme case of disfluent language: sentences from the original corpus where the words are reordered randomly. We do this on the source or target side. 

\paragraph{\textsc{Wrong Language}}
A parallel corpus may be polluted by text in a third language, say French in a German--English corpus. 
This may occur on the source or target side of the parallel corpus. To simulate this, we add French--English (bad source) or German--French (bad target) data to a German--English corpus.

\paragraph{\textsc{Untranslated Sentences}}
Especially in parallel corpora crawled from the web, there are often sentences that are untranslated from the source in the target. Examples are navigational elements or copyright notices in the footer. Purportedly multi-lingual web sites may be only partially translated, while some original text is copied. 
Again, this may show up on the source or the target side. We take sentences from either the source or target side of the original parallel corpus and simply copy them to the other side.

\paragraph{\textsc{Short Segments}}
Sometimes additional data comes in the form of bilingual dictionaries. Can we simply add them as additional sentence pairs, even if they consist of single words or short phrases? 
We simulate this kind of data by sub-subsampling a parallel corpus to include only sentences of maximum length 2 or 5.

\section{Experimental Setup}

\subsection{Neural Machine Translation}
Our neural machine translation systems are trained using Marian \citep{mariannmt}.\footnote{\href{ https://marian-nmt.github.io/}{marian-nmt.github.io}} We build shallow RNN-based encoder-decoder models with attention \citep{bahdanau:ICLR:2015}.
We train Byte-Pair Encoding segmentation models (BPE) \cite{bpe} with a vocab size of $50,000$ on both sides of the parallel corpus for each experiment. 
We apply drop-out with $20\%$ probability on the RNNs, and with $10\%$ probability on the source and target words.
We stop training after convergence of cross-entropy on the development set, and we average the 4 highest performing models (as determined by development set BLEU performance) to use as an ensemble for decoding (checkpoint assembling). 
Training of each system takes 2--4 days on a single GPU (GTX 1080ti).

While we focus on RNN-based models with attention as our NMT architecture, we note that different architectures have been proposed,  including based on convolutional neural networks \citep{kalchbrenner-blunsom:2013:EMNLP, gehring-EtAl:2017:Long} and the self-attention based Transformer model \citep{DBLP:journals/corr/VaswaniSPUJGKP17}. 

\subsection{Statistical Machine Translation}
Our statistical machine translation systems are trained using Moses \citep{koehn-EtAl:2007:PosterDemo}.\footnote{\href {http://www.statmt.org/moses/}{statmt.org/moses}} We build phrase-based systems using standard features commonly used in recent system submissions to WMT \citep{haddow-EtAl:2015:WMT,ding-EtAl:2016:WMT,ding-EtAl:2017:WMT}.
We trained our systems with the following settings: a maximum sentence length of 80, grow-diag-final-and symmetrization of GIZA++ alignments, an interpolated Kneser-Ney smoothed 5-gram language model with KenLM \cite{heafield:2011:WMT}, hierarchical lexicalized reordering \cite{galley-manning:2008:EMNLP}, a lexically-driven 5-gram operation sequence model (OSM) \cite{durrani-EtAl:2013:ACL}, sparse domain indicator, phrase length, and count bin features \cite{blunsom-osborne:2008:EMNLP,chiang-knight-wang:2009:NAACLHLT09}, a maximum phrase-length of 5, compact phrase table \cite{EAMT-2012-Junczys-Dowmunt} minimum Bayes risk decoding \cite{KumarB04}, cube pruning \cite{huang-chiang:2007:ACLMain}, with a stack-size of 1000 during tuning. We optimize feature function weights with k-best MIRA \cite{cherry-foster:2012:NAACL-HLT}.

While we focus on phrase based systems as our SMT paradigm, we note that there are other statistical machine translation approaches such as hierarchical phrase-based models \citep{Chiang:CL:2007} and syntax-based models \citep{Galley:2004,galley-EtAl:2006:COLACL} that may have better performance in certain language pairs and in low resource conditions. 

\subsection{Clean Corpus}
In our experiments, we translate from German to English. We use datasets from the shared translation task organized
alongside the Conference on Machine Translation
(WMT)\footnote{\href {http://www.statmt.org/wmt17/}{statmt.org/wmt17/}} as clean training data. For our baseline we use: Europarl \cite{Koehn:2005:MTS},\footnote{\href{http://www.statmt.org/europarl/}{statmt.org/europarl}} News Commentary,\footnote{\href{http://www.casmacat.eu/corpus/news-commentary.html}{casmacat.eu/corpus/news-commentary.html}} and the Rapid EU Press Release parallel corpus. The corpus size is about 83 million tokens per language. We use {\tt{newstest2015}} for tuning SMT systems,  {\tt{newstest2016}} as a development set for  NMT systems, and report results on {\tt{newstest2017}}.

Note that we do not add monolingual data to our systems since this would make our study more complex. So, we always train our language model on the target side of the parallel corpus for that experiment. While using monolingual data for language modelling  is standard practice in statistical machine translation, how to use such data for neural models is less obvious.

\subsection{Noisy Corpora}
For \textsc{misaligned sentence} and \textsc{misordered word} noise, we use the clean corpus (above) and perturb the data. To create \textsc{untranslated sentence} noise, we also use the clean corpus and create pairs of identical sentences.

For \textsc{wrong language} noise, we do not have French--English and German--French data of the same size. Hence, we use the EU Bookstore corpus \citep{SkadinsEA:LREC14}.\footnote{\href{http://opus.nlpl.eu/EUbookshop.php}{opus.nlpl.eu/EUbookshop.php}}

The \textsc{short segments} are extracted from OPUS corpora \cite{Tiedemann:RANLP5, opus,opensubtitles2016}:\footnote{\href{ http://opus.nlpl.eu/}{opus.nlpl.eu}}  EMEA (descriptions of medicines),\footnote{\href{http://www.emea.europa.eu/}{emea.europa.eu}}  Tanzil (religious text),\footnote{\href{http://tanzil.net/trans/}{tanzil.net/trans}}  Open Subtitles 2016,\footnote{\href{ http://www.opensubtitles.org/}{opensubtitles.org} } Acquis (legislative text),\footnote{\href{ https://ec.europa.eu/jrc/en/language-technologies/jrc-acquis }{ec.europa.eu/jrc/en/language-technologies/jrc-acquis}} GNOME (software localization files),\footnote{\href{https://l10n.gnome.org}{l10n.gnome.org}} KDE (localization files), PHP (technical manual),\footnote{\href{http://se.php.net/download-docs.php}{se.php.net/download-docs}} Ubuntu (localization files),\footnote{\href{ https://translations.launchpad.net}{translations.launchpad.net}} and Open Office.\footnote{\href{ http://www.openoffice.org/ }{openoffice.org}}  We use only pairs where both the English and German segments are at most 2 or 5 words long. Since this results in small data sets (2 million and 15 tokens per language, respectively), they are duplicated multiple times.

We also show the results for naturally occurring noisy web data from the raw 2016 ParaCrawl corpus (deduplicated raw set).\footnote{\href{https://paracrawl.eu/}{paracrawl.eu}}

We sample the noisy corpus in an amount equal to $5\%$, $10\%$, $20\%$, $50\%$, and $100\%$ of the clean corpus. This reflects the realistic situation where there is a clean corpus, and one would like to add additional data that has the potential to be noisy. 
For each experiment, we use the target side of the parallel corpus to train the SMT language model, including the noisy text.

\section{Impact on Translation Quality} 

Table \ref{tab:results} shows the effect of adding each type of noise to the clean corpus.\footnote{We report case-sensitive detokenized
BLEU \cite{papineni-EtAl:2002:ACL} calculated using
mteval-v13a.pl.} For some types of noise NMT is harmed more than SMT: \textsc{mismatched sentences} (up to -1.9 for NMT, -0.6 for SMT), \textsc{misordered words} (source) (-1.7 vs. -0.3), \textsc{wrong language} (target) (-2.2 vs. -0.6).

\textsc{ Short segments}, \textsc{ untranslated source sentences} and \textsc{ wrong source language} have little impact on either (at most a degradation of -0.7). \textsc{ Misordered target words} decreases BLEU scores for both SMT and NMT by just over 1 point (100\% noise).

The most dramatic difference is \textsc{ untranslated target sentence} noise. When added at 5\% of the original data, it degrades NMT performance by $9.6$  BLEU, from $27.2$ to  $17.6$. Adding this noise at 100\% of the original data degrades performance by $24.0$ BLEU, dropping the score from  $27.2$ to  $3.2$. 
In contrast, the SMT system only drops $2.9$ BLEU, from $24.0$ to $21.1$.

\newcommand{\debugcolor}{white}

\newcommand{\barchart}[5]{\begin{tikzpicture}[scale=0.15]
\fill[\debugcolor] (5.7,0) rectangle (5.8,-#1);
\fill[\debugcolor] (5.7,0) rectangle (5.8,4);
\fill[green] (0,0) rectangle (5.5,0#3);
\fill[blue] (6,0) rectangle (11.5,0#5);
\draw (2.75,0) node[anchor=south] {#2};
\draw (8.75,0) node[anchor=south] {#4};
\draw (2.75,0#3) node[anchor=north] {\textcolor{red}{#3}};
\draw (8.75,0#5) node[anchor=north] {\textcolor{red}{#5}};
\end{tikzpicture}}

\newcommand{\positivebarchart}[5]{\begin{tikzpicture}[scale=0.15]
\fill[\debugcolor] (5.7,0) rectangle (5.8,-#1);
\fill[\debugcolor] (5.7,0) rectangle (5.8,4);
\fill[green] (0,0) rectangle (5.5,0#3);
\fill[blue] (6,0) rectangle (11.5,0#5);
\draw (2.75,0) node[anchor=south] {#2};
\draw (8.75,0) node[anchor=south] {#4};
\draw (2.75,0) node[anchor=north] {\textcolor{darkgreen}{#3}};
\draw (8.75,0) node[anchor=north] {\textcolor{darkgreen}{#5}};
\end{tikzpicture}}

\newcommand{\negposbarchart}[5]{\begin{tikzpicture}[scale=0.15]
\fill[\debugcolor] (5.7,0) rectangle (5.8,-#1);
\fill[\debugcolor] (5.7,0) rectangle (5.8,4);
\fill[green] (0,0) rectangle (5.5,0#3);
\fill[blue] (6,0) rectangle (11.5,0#5);
\draw (2.75,0) node[anchor=south] {#2};
\draw (8.75,0) node[anchor=south] {#4};
\draw (2.75,0#3) node[anchor=north] {\textcolor{red}{#3}};
\draw (8.75,0) node[anchor=north] {\textcolor{darkgreen}{#5}};
\end{tikzpicture}}

\newcommand{\rowlabel}[2]{\begin{tikzpicture}[scale=0.15]
\fill[\debugcolor] (0,0) rectangle (0.1,4);
\fill[\debugcolor] (0,0) rectangle (0.1,-#1);
\draw (0,0) node[anchor=south] {\bf #2}; 
\end{tikzpicture}}
\newcommand{\rowlabelx}[3]{\begin{tikzpicture}[scale=0.15]
\fill[\debugcolor] (0,0) rectangle (0.1,4);
\fill[\debugcolor] (0,0) rectangle (0.1,-#1);
\draw (0,0) node[anchor=south] {\bf #2}; 
\draw (1.4,-3.5) node[anchor=south] {\bf #3}; 
\end{tikzpicture}}

\begin{table*}
\addtolength{\tabcolsep}{-2pt} 
\begin{tabular}{l|c|c|c|c|c|}
 & \bf 5\% & \bf 10\%& \bf 20\%& \bf 50\%& \bf 100\% \\ \hline
\rowlabel{5.5}{\textsc{Misaligned sentences}} & 
    \barchart{5.5}{26.5}{-0.7}{24.0}{-0.0} & 
    \barchart{5.5}{26.5}{-0.7}{24.0}{-0.0} & 
    \barchart{5.5}{26.3}{-0.9}{23.9}{-0.1} & 
    \barchart{5.5}{26.1}{-1.1}{23.9}{-0.1} & 
    \barchart{5.5}{25.3}{-1.9}{23.4}{-0.6}\\\hline
\rowlabelx{5.5}{\textsc{Misordered words}}{\textsc{(source)}} & 
    \barchart{5.5}{26.9}{-0.3}{24.0}{-0.0} & 
    \barchart{5.5}{26.6}{-0.6}{23.6}{-0.4} & 
    \barchart{5.5}{26.4}{-0.8}{23.9}{-0.1} & 
    \barchart{5.5}{26.6}{-0.6}{23.6}{-0.4} &
    \barchart{5.5}{25.5}{-1.7}{23.7}{-0.3}\\
\rowlabelx{5.5}{\textsc{Misordered words}}{\textsc{(target)}} & 
    \barchart{5.5}{27.0}{-0.2}{24.0}{-0.0} & 
    \barchart{5.5}{26.8}{-0.4}{24.0}{-0.0} & 
    \barchart{5.5}{26.4}{-0.8}{23.4}{-0.6} & 
    \barchart{5.5}{26.7}{-0.5}{23.2}{-0.8} &
    \barchart{5.5}{26.1}{-1.1}{22.9}{-1.1}\\\hline
\rowlabelx{5.5}{\textsc{Wrong language}}{\textsc{(French source)}} & 
    \barchart{5.5}{26.9}{-0.3}{24.0}{-0.0} & 
    \barchart{5.5}{26.8}{-0.4}{23.9}{-0.1} & 
    \barchart{5.5}{26.8}{-0.4}{23.9}{-0.1} & 
    \barchart{5.5}{26.8}{-0.4}{23.9}{-0.1} &
    \barchart{5.5}{26.8}{-0.4}{23.8}{-0.2}\\
\rowlabelx{5.5}{\textsc{Wrong language}}{\textsc{(French target)}} & 
    \barchart{5.5}{26.7}{-0.5}{24.0}{-0.0} & 
    \barchart{5.5}{26.6}{-0.6}{23.9}{-0.1} & 
    \barchart{5.5}{26.7}{-0.5}{23.8}{-0.2} & 
    \barchart{5.5}{26.2}{-1.0}{23.5}{-0.5} &
    \barchart{5.5}{25.0}{-2.2}{23.4}{-0.6}\\ \hline
\rowlabelx{5.5}{\textsc{Untranslated }}{\textsc{(English source)}}  & 
    \barchart{5.5}{27.2}{-0.0}{23.9}{-0.1} & 
    \barchart{5.5}{27.0}{-0.2}{23.9}{-0.1} & 
    \barchart{5.5}{26.7}{-0.5}{23.6}{-0.4} & 
    \barchart{5.5}{26.8}{-0.4}{23.7}{-0.3} &
    \barchart{5.5}{26.9}{-0.3}{23.5}{-0.5}\\ 
\rowlabelx{28}{\textsc{Untranslated }}{\textsc{(German target)}}  & 
    \barchart{28}{17.6}{ -9.8}{23.8}{-0.2} & 
    \barchart{28}{11.2}{-16.0}{23.9}{-0.1} & 
    \barchart{28}{ 5.6}{-21.6}{23.8}{-0.2} & 
    \barchart{28}{ 3.2}{-24.0}{23.4}{-0.6} & 
    \barchart{28}{ 3.2}{-24.0}{21.1}{-2.9}\\ \hline
\rowlabelx{5.5}{\textsc{Short segments}}{(max 2)} & 
    \negposbarchart{5.5}{27.1}{-0.1}{24.1}{+0.1} & 
    \barchart{5.5}{26.5}{-0.7}{23.9}{-0.1} & 
    \barchart{5.5}{26.7}{-0.5}{23.8}{-0.2} &&\\
\rowlabelx{5.5}{\textsc{Short segments}}{(max 5)} & 
    \positivebarchart{5.5}{27.8}{+0.6}{24.2}{+0.2} & 
    \positivebarchart{5.5}{27.6}{+0.4}{24.5}{+0.5} & 
    \positivebarchart{5.5}{28.0}{+0.8}{24.5}{+0.5} & 
    \negposbarchart{5.5}{26.6}{-0.6}{24.2}{+0.2} &\\ \hline
\rowlabel{13.5}{\textsc{Raw crawl data}} &
    \positivebarchart{13.5}{27.4}{+0.2}{24.2}{+0.2} & 
    \negposbarchart{13.5}{26.6}{-0.6}{24.2}{+0.2} &
    \negposbarchart{13.5}{24.7}{-2.5}{24.4}{+0.4} &
    \negposbarchart{13.5}{20.9}{-6.3}{24.8}{+0.8} &
    \negposbarchart{13.5}{17.3}{-9.9}{25.2}{+1.2}\\ \hline
\end{tabular}
\caption{Results from adding different amounts of noise (ratio of original clean corpus) for various types of noise in German-English Translation. Generally neural machine translation (left green bars) is harmed more than statistical machine translation (right blue bars). The worst type of noise are segments in the source language copied untranslated into the target.}
\label{tab:results}
\end{table*}

\subsection{Copied output}
Since the noise type where the target side is  a copy of the source has such a big impact, we examine the system output in more detail.

\begin{figure}
\hspace{-11pt}
\includegraphics{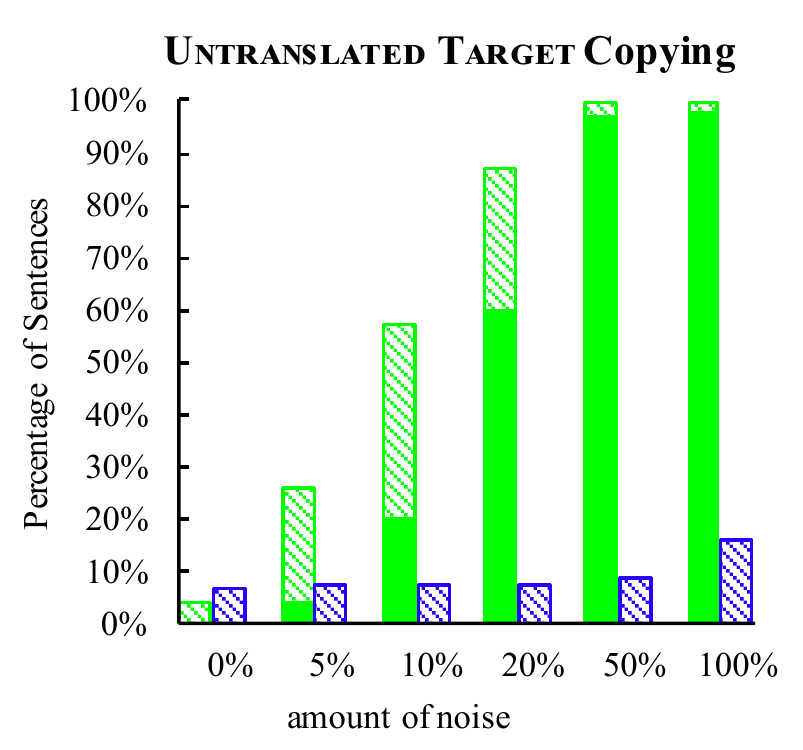}
\vspace{-25pt}
\caption{Copied sentences in the \textsc{Untranslated (target)} experiments. NMT is the left green bars, SMT is the right blue bars. Sentences that are exact matches to the source are the solid bars, sentences that are more similar to the source than the target are the shaded bars.}
\label{fig:Untanslated_copy}
\end{figure}

\begin{figure}
\hspace{-11pt}
\includegraphics{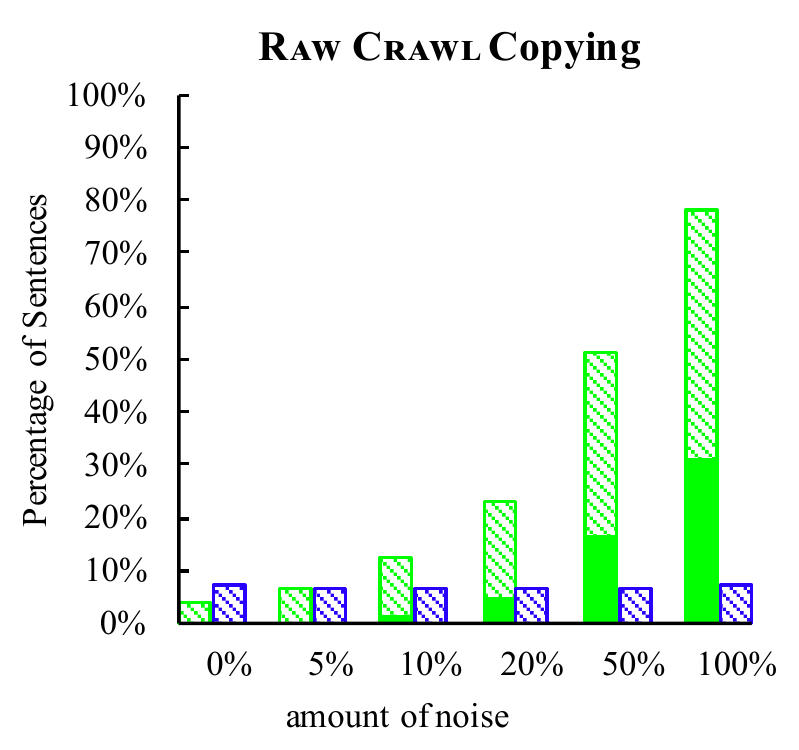}
\vspace{-25pt}
\caption{Copied sentences in the \textsc{raw crawl} experiments. NMT is the left green bars, SMT is the right blue bars. Sentences that are exact matches to the source are the solid bars, sentences that are more similar to the source than the target are the shaded bars.}
\label{fig:Untanslated_paracrawl}
\end{figure}

We report the percent of sentences in the evaluation set that are identical to the source for the \textsc{untranslated target sentence} and \textsc{ raw crawl} data in Figures~\ref{fig:Untanslated_copy} and \ref{fig:Untanslated_paracrawl} (solid bars). The SMT systems output 0 or 1 sentences that are  exact copies. However, with just $20\%$ of the \textsc{untranslated target sentence} noise, $60\%$ of the NMT output sentences are identical to the source. 

This suggests that the NMT systems learn to copy, which may be useful for named entities. However, with even a small amount of this data it is doing far more harm than good.

 Figures~\ref{fig:Untanslated_copy} and \ref{fig:Untanslated_paracrawl} show the percent of sentences that have a  worse TER score against the reference than against the 
source (shaded bars). 
This means that it would take fewer edits to transform the sentence into the source than it would to transform it into the target. 
When just 10\% \textsc{untranslated target sentence} data is added, 57\% of the sentences are more similar to the source than to the reference, indicating partial copying.

\begin{figure}
\hspace{-7pt}
\includegraphics[width=3.1in]{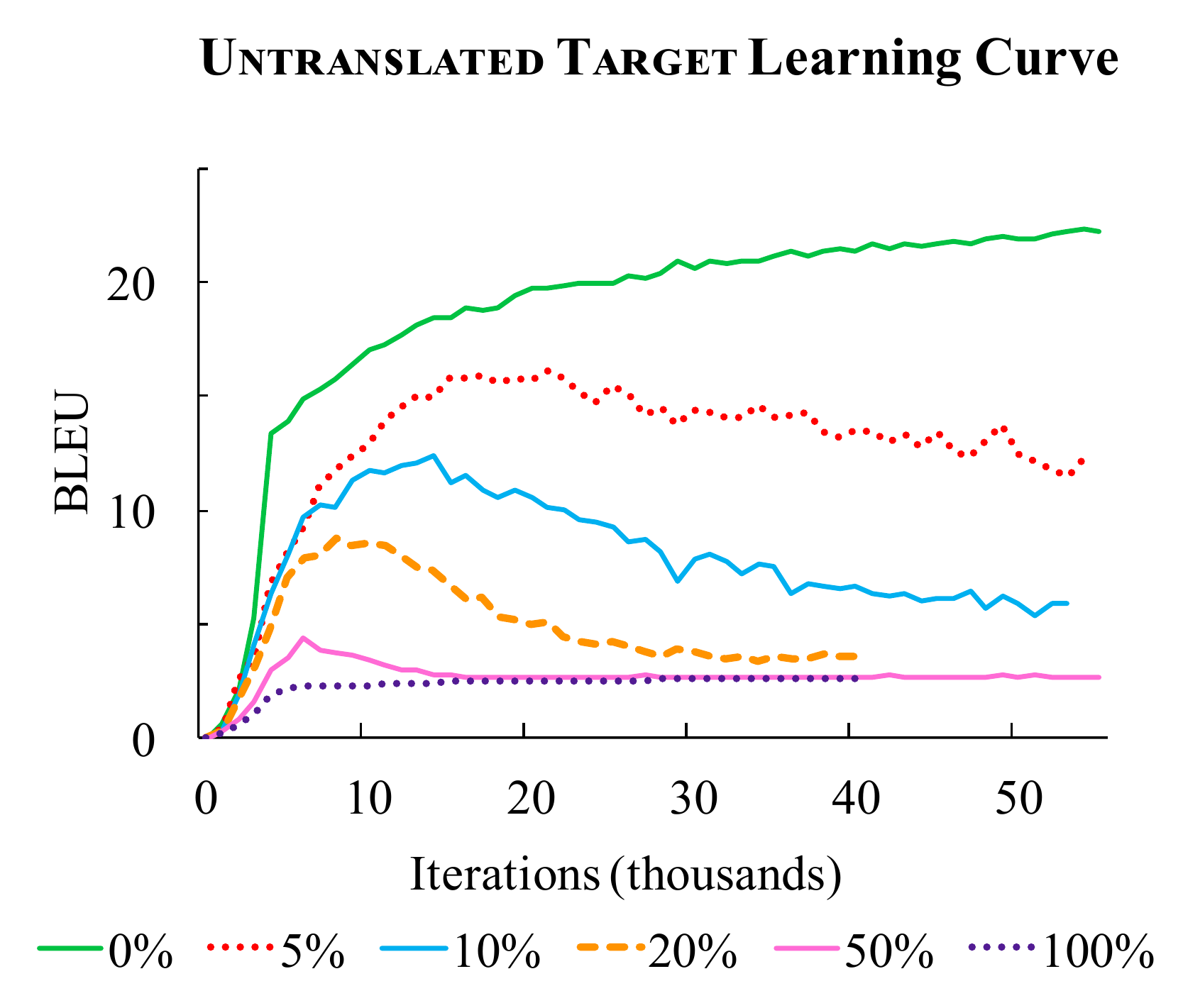}
\caption{Learning curves for the NMT \textsc{ untranslated target sentence} experiments. 
}
\label{fig:learningcurve}
\vspace{-15pt}
\end{figure}

This suggests that the NMT system is overfitting the copied portion of the training corpus. This is supported by Figure \ref{fig:learningcurve}, which shows the learning curve on the development set for the \textsc{untranslated target sentence} noise setup. The performance for the systems trained on noisy corpora begin to improve, before over-fitting to the copy portion of the training set. Note that we plot the BLEU performance on the development set with beam search, while the system is optimizing cross-entropy given a perfect prefix.

Other work has also considered copying in NMT. \citet{currey-micelibarone-heafield:2017:WMT} add copied data and back-translated data to a clean parallel corpus. They report improvements on EN $\leftrightarrow$ RO when adding as much back-translated and copied data as they have parallel (1:1:1 ratio). For EN$\leftrightarrow$TR and EN$\leftrightarrow$DE, they add twice as much back translated and copied data as parallel data (1:2:2 ratio), and report improvements on EN$\leftrightarrow$TR but not on  EN$\leftrightarrow$DE. However, their EN$\leftrightarrow$DE
systems trained with the copied corpus did not
perform  worse than baseline systems. 
\citet{DBLP:journals/corr/abs-1803-00047} found that while copied training sentences represent less than $2.0\%$ of their training data (WMT 14 EN$\leftrightarrow$DE and EN$\leftrightarrow$FR), copies are over-represented in the output of beam search. Using a subset of training data from WMT 17, they replace a subset of the true translations with a copy of the input. They analyze varying amounts of copied noise, and a variety of beam sizes. Larger beams are more effected by this kind of noise; however, for all beam sizes performance degrades completely with $50\%$ copied sentences.\footnote{See Figure 3 in \citet{DBLP:journals/corr/abs-1803-00047}.} 

\subsection{Incorrect Language output}
Another interesting case is when a German--French corpus is added to a German--English corpus ({\textsc{wrong target language}). Both neural and statistical machine translation are surprisingly robust, even when these corpora are provided in equal amounts.

We performed a manual analysis of the neural machine translation experiments. 
For the each of the  noise levels, we report the percentage of NMT output sentences in French (out of of $3004$: $5\%$: $0.20\%$, $10\%$: $0.60\%$, $20\%$: $1.7\%$, $50\%$: $3.3\%$, $100\%$: $6.7\%$.
Most NMT output sentences were either entirely French or English, with the exception of a few mis-translated cognates (e.g.: `fa{\c{c}}ade', `accessibilit\'e').

In the SMT experiment with $100\%$ noisy data added, there are a couple of French words in mostly English sentences. These are much less frequent than unknown German words passed through.  Only 1 sentence is mostly French.

It is surprising that such a small percentage of the output sentences were French, since up to half of the target data in training was in French. 
We attribute this to the domain of the added data differing from the test data.
Source sentences in the test set are more similar to the domain-relevant clean parallel training corpus than the domain-divergent noise corpus.

\section{Conclusion} 
We defined five types of noise in parallel data, motivated by a study of raw web crawl data. We found that neural machine translation is less robust to many types of noise than statistical machine translation. In the most extreme case, when the reference is an untranslated copy of the source data, neural machine translation may learn to excessively copy the input. These findings should inform future work on corpus cleaning.

\section*{Acknowledgements}
This work has been partially supported by the DARPA LORELEI and the IARPA MATERIAL programs. It also benefited from the Paracrawl web crawling project which is funded by a Google faculty grant and the European Union (CEF).
 
\bibliographystyle{acl_natbib}
\bibliography{mt,more,ds,naaclhlt}

\end{document}